\documentclass{ieeeaccess}
\usepackage{cite}
\usepackage{amsmath,amssymb,amsfonts}
\usepackage{algorithmic}
\usepackage{blindtext}
\usepackage{hyperref}
\usepackage{graphicx}
\usepackage{url}
\usepackage{textcomp}
\usepackage{multirow}
\def\BibTeX{{\rm B\kern-.05em{\sc i\kern-.025em b}\kern-.08em
    T\kern-.1667em\lower.7ex\hbox{E}\kern-.125emX}}
\begin{document}
\history{Date of publication xxxx 00, 0000, date of current version xxxx 00, 0000.}
\doi{10.1109/ACCESS.2017.DOI}


\title{Autonomous and Human-Driven Vehicles Interacting in a Roundabout: A Quantitative and Qualitative Evaluation}

\author{\uppercase{Laura Ferrarotti$^{*}$\authorrefmark{2}, Massimiliano Luca$^{*}$ \authorrefmark{1,2} , Gabriele Santin\authorrefmark{2,6},  Giorgio Previati\authorrefmark{3}, Gianpiero Mastinu\authorrefmark{3}, Massimiliano Gobbi\authorrefmark{3}, Elena Campi\authorrefmark{3}, Lorenzo Uccello\authorrefmark{3}, Antonino Albanese\authorrefmark{4}, 
Praveen Zalaya\authorrefmark{5}, Alessandro Roccasalva\authorrefmark{5}, Bruno Lepri\authorrefmark{2}}}

\address[1]{Free University of Bolzano, Faculty of Computer Science, Bolzano, Italy}
\address[2]{Fondazione Bruno Kessler, Mobile and Social Computing Laboratory, Trento, Italy}
\address[3]{Politecnico di Milano, Department of Mechanical Engineering, Milano, Italy}
\address[4]{Italtel S.p.A., Milano, Italy}
\address[5]{Centro Ricerche Fiat, TECH, Orbassano, Italy}
\address[6]{Ca' Foscari University of Venice, Venice, Italy}

\tfootnote{$^{*}$ The authors contributed equally to the work. Alphabetically sorted.}

\markboth
{Ferrarotti, Luca, \headeretal: Autonomous and Human-Driven Vehicles Interacting in a Roundabout}
{Ferrarotti, Luca, \headeretal: Autonomous and Human-Driven Vehicles Interacting in a Roundabout}

\corresp{Corresponding author: Massimiliano Luca (e-mail: mluca@fbk.eu)}

\begin{abstract}
Optimizing traffic dynamics in an evolving transportation landscape is crucial, particularly in scenarios where autonomous vehicles (AVs) with varying levels of autonomy coexist with human-driven cars. {\color{black}While optimizing Reinforcement Learning (RL) policies for such scenarios is becoming more and more common, little has been said about realistic evaluations of such trained policies.} This paper presents {\color{black}an evaluation of the effects of AVs penetration among human drivers in a roundabout scenario, considering both quantitative and qualitative aspects. In particular, we} 
learn a policy to minimize traffic jams (i.e., minimize the time to cross the scenario) and to minimize pollution in a roundabout in Milan, Italy. Through empirical analysis, we demonstrate that 
{\color{black} the presence of AVs} can reduce time and pollution levels. Furthermore, we qualitatively evaluate the learned policy using a cutting-edge cockpit to assess its performance in near-real-world conditions. To gauge the practicality and acceptability of the policy, we conduct evaluations with human participants using the simulator, focusing on a range of metrics like traffic smoothness and safety perception. In general, our findings show that human-driven vehicles benefit from optimizing AVs dynamics. Also, participants in the study highlight that the scenario with 80\% AVs is perceived as safer than the scenario with 20\%. The same result is obtained for traffic smoothness perception. 

\end{abstract}

\begin{keywords}
Transportation, Autonomous Vehicles, Urban Mobility, Reinforcement Learning
\end{keywords}

\titlepgskip=-15pt

\maketitle

\section{Introduction}
\label{sec:introduction}
Modern society grapples with a large amount of societal challenges. Among the most pressing is the constant increase in levels of pollution and related traffic congestions \cite{united2022sustainable}, both of which threaten the sustainability and livability of our urban environments \cite{xie2017review, luca2023crime}. At the same time, at the forefront of the technological revolution are autonomous vehicles (AVs). These vehicles, driven by advanced algorithms and sensors, promise to redefine how we perceive transportation \cite{thomas2020perception,chikaraishi2020risk,hilgarter2020public}. 
They can potentially alleviate some of the most persistent issues of modern urban transportation, from increasing safety \cite{wang2020safety,mariani2018overview,koopman2019safety,koopman2017autonomous,ye2019evaluating} to optimizing traffic flow \cite{mushtaq2021traffic,zambrano2019centralized,namazi2019intelligent,miglani2019deep}. However, as with any nascent technology, the real-world implementation of AVs is laden with challenges. Among the most significant are the safety, and prohibitive costs associated with testing and validating the efficiency of these vehicles in real conditions. {\color{black} One aspect that is fundamental to address these challenges is the evaluation of driving scenarios in which AVs are present together with human drivers, in order to assess not only the performance attained by the AVs, but the behavior of the whole hybrid multi-agent system of drivers, together with the perception of driving comfort and safety experienced by the human drivers involved.}
In our paper, {\color{black} in order to conduct such analysis conjugating both safety and realistic conditions,} we bypass the logistical and financial constraints of real-world testing by harnessing the power of state-of-the-art simulation tools. Our primary focus is on a small-scale yet intricately complex scenario: a roundabout in Milan, Italy. Utilizing Simulation of Urban MObility (SUMO) \cite{krajzewicz2002sumo}, a cutting-edge traffic simulator, we create a realistic environment where both AVs and human-driven vehicles (HVs) coexist, navigating the roundabout under realistic traffic loads. Also, we bridge the gap between static simulation and real-world experience by integrating SUMO with VI-WorldSim\footnote{\href{https://www.vi-grade.com/en/products/vi-worldsim/}{https://www.vi-grade.com/en/products/vi-worldsim/}} \cite{previati2023sumo}. This user-friendly, fully integrated graphic environment not only accelerates vehicle development offline but also facilitates a more immersive experience on driving simulators. To enhance the realism of our study, we leverage a high-fidelity cockpit that replicates real-world driving conditions installed at the DrisMi Lab at the Polytechnic University of Milan\footnote{\href{https://www.drismi.polimi.it/drismi/}{https://www.drismi.polimi.it/drismi/}}, enabling us to evaluate, to the best of our knowledge for the first time, the scenario from a qualitative perspective. In our simulation, HVs adhere to realistic dynamics as simulated by SUMO, while AVs actions are dictated by a policy learned via Reinforcement Learning (RL). 

{In general, the integration of microscopic traffic simulators with driving simulators is not a novel concept \cite{espie2006joint, vladisavljevic2009importance}. However, previous studies have presented certain challenges, which our paper addresses and resolves \cite{hasan2021distributed,barthauer2019testing,barthauer2018coupling,kaths2019co}. Specifically, our approach ensures \emph{synchronized} co-simulation in \emph{real-time}, utilizing \emph{the same road network} in both simulators and achieving an impressively low delay of 5 ms.}

Our findings, as detailed in this paper, shed light on the multifaceted impact of AV integration. While the benefits of AVs in reducing pollution and alleviating congestion are evident, interestingly enough, the presence of AVs also augments the efficiency of human-driven vehicles. 
{\color{black}As we show in the Tables in Section \ref{sec:ql_eval}, as the AV penetration rate increases, the ripple effects are felt across the entire traffic ecosystem, offering insights into a future where harmonious coexistence between AVs and HVs might redefine urban transportation. More precisely, we evaluate the learned policy and the interaction between AVs and HVs qualitatively using traffic smoothness perception and safety perception as metrics. The evaluation is carried by surveying the perception of individuals who use the cockpit installed at the DrisMi Lab. While a number of participants felt a difference between the presence of 20\% and 80\% of AVs on the streets, most of the people preferred the scenario with an 80\% penetration rate of AVs both in terms of safety and smoothness as highlighted in the Tables of Section V.B.}

The advances presented in our study can be summarized as follows: 

\begin{itemize}
    \item We propose a framework that consists of the integration of three realistic simulators: SUMO, VI-WorldSim, and a cockpit. This framework enables quantitative and qualitative evaluation of AVs and HVs interactions. 
    \item We {\color{black} employ} RL to learn AVs behaviors in a real-world scenario (i.e., a roundabout in Milan, Italy) with realistic traffic loads.
    \item We measure the reduction of crossing time (up to -10,72\% for AVs and -8,52\% for HVs), emissions (up to -38,98\% for AVs and -39,13\% for HVs) and consumption (up to -35,82\% for AVs and -35,15\% for HVs) as a quantitative metric for policy evaluation, and traffic smoothness perception and safety perception as qualitative metrics.
\end{itemize}

The rest of the paper is structured as follows. In Section \ref{sec:back}, we briefly introduce SUMO, VI-WorldSim, the cockpit and RL concepts as background notions. We then conduct a literature review of works that study mixed-traffic scenarios, and of works leveraging RL for AV and traffic simulation in Section \ref{sec:litrev}. In Section \ref{sec:methodology}, we first discuss how we turn SUMO into an RL environment, the algorithm we use, namely {\color{black}Proximal Policy Optimization (PPO) \cite{shulman2017proximal}}, how the test environment (i.e., the roundabout in Milan) is designed, and how the data for policy fine-tuning and traffic loads are collected. In Section \ref{sec:res}, we present both the quantitative and qualitative results, and we comment on some of the findings. In Section \ref{sec:conclusion}, we summarize the paper and present some interesting future directions to follow.

\section{Background}
\label{sec:back}
\subsection{Simulations in SUMO}
{Simulations are extensively used in a variety of fields such as urban planning \cite{waddell2002urbansim, mauro2022generating}, transportation \cite{barcelo2010fundamentals,pursula1999simulation,khaidem2020optimizing}, robotics \cite{todorov2012mujoco}, epidemiology \cite{eubank2004modelling, cencetti2021digital}, gaming \cite{brockman2016openai}, and others. Experiments in the real world are often costly, dangerous, and infeasible. Thus, simulators provide a solution for evaluating hypotheses and methodologies \emph{in silico} where certain aspects of the behavior faithfully mirror the real world. 

SUMO \cite{krajzewicz2002sumo} is a state-of-the-art multi-agent simulator for transportation systems that reproduce realistic behaviors of drivers. In SUMO, it is possible to deploy multiple agents that use different transportation means (e.g., cars, public transit, bicycles) to reach different goals. The agents can move within a street network that defines the environment in which they can operate. Interestingly enough, SUMO is also a \emph{microscopic} traffic simulator, i.e., each agent is modeled as an individual based on separate and different car-following and lane-changing models. 

SUMO's workflow to simulate realistic traffic is organized as follows. First, the road network is defined to match the real world, and road loads (e.g., a realistic number of agents using a specific street) are specified.  Second, the agents are executed in the road network through high-fidelity simulations and scored according to a cost function that measures if certain goals have been reached. Next, as in this work we 
use a learned {\color{black} Reinforcement Learning policy trained in collaboration among all the AVs involved}, SUMO re-plans, re-executes, and re-scores the actions taken by the agents using a co-evolutionary algorithm until nobody can unilaterally improve their trips. At that point, the system has an equilibrium, and we can inspect the individuals' typical behaviors.}

\subsection{VI-WorldSim}
\label{sec:wsim}
{ 
VI-WorldSim is an innovative software solution designed to facilitate creating and testing lifelike driving scenarios. These scenarios include traffic flow, pedestrians, weather conditions, and sensor feedback. 
By utilizing VI-WorldSim, individuals can immerse themselves in many realistic environments, from bustling urban streets to expansive highways and specialized test sites. This versatility allows for an in-depth evaluation of how a vehicle model responds to many situations, ensuring that every potential challenge is addressed during development. Also, the realism of the simulations combined with the integration with realistic cockpits can be used to evaluate scenarios qualitatively safely and relatively cheaply. An example of what a VI-WorldSim simulation looks like can be seen in Figure \ref{fig:simulator} B. 

{\color{black} \subsection{VI-Worldsim and SUMO Interaction}
The traffic scenario is generated thanks to a co-simulation between SUMO, which is used as a traffic engine, and VI-WorldSim, which simulates the vehicle's motion driven by the human in the loop and allows a graphical representation of the traffic scenario. SUMO is in charge of simulating all virtual vehicles in the roundabout, comprising HVs and AVs. In particular, SUMO receives the data about the car driven by the human in the loop (ego-vehicle) from the driving simulator through VI-WorldSim. The current position of all simulated vehicles is fed to VI-WorldSim which is in charge of all the graphical environment of the driving simulator, of the interface with the human in the loop and of the simulation of the motion of the ego-vehicle according to the request of the real human driver. All these simulations are performed in real-time and the corresponding data are stored in a real-time database.}
{Additional information about the communication schema between SUMO and VI-WorldSim can be found in \cite{previati2023sumo}}
}
\subsection{The Cockpit}
\label{sec:cpit}
\begin{figure}[ht!]
    \centering
    \includegraphics[width=1\columnwidth]{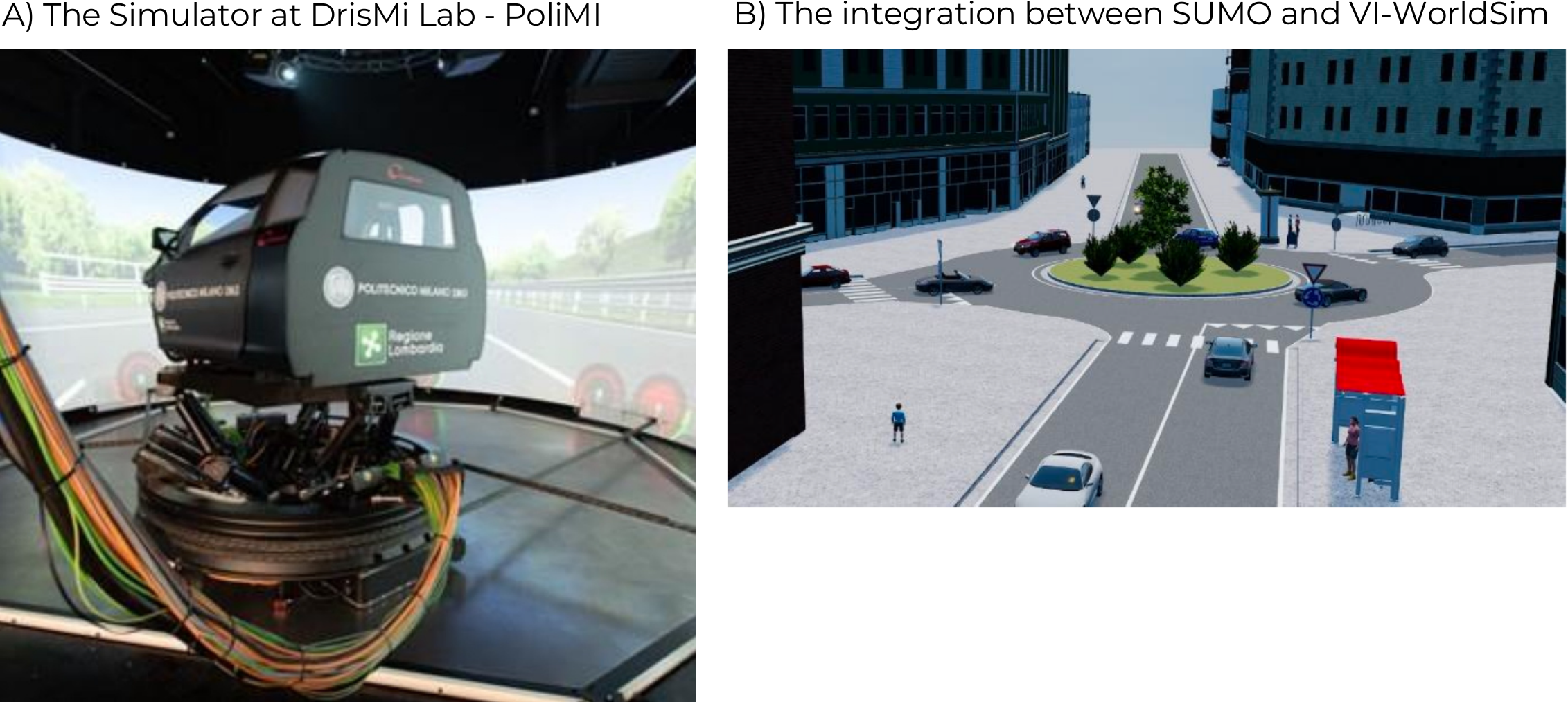}
    \caption{\textbf{A)} The cockpit was installed at DrisMi Laboratory of the Polytechnic University of Milan. \textbf{B)} An example of the output of VI-WorldSim after the integration with SUMO.}
    \label{fig:simulator}
\end{figure}

{

The human in the loop drives the vehicle from the cockpit of the driving simulator, which moves accordingly to the simulated motion of the ego vehicle. The cockpit can be seen in Figure \ref{fig:simulator} A. The cockpit is equipped with a telematic box, which is connected to the Controller Area Network bus of the cockpit, reads the dynamic data of the car, and transmits them to a 5G radio platform. The 5G radio platform transmits the data to and from an edge server and is controlled by a Next Unit of Computing (NUC) where the services (e.g., learned policy) are installed. Finally, the edge server hosts the RL infrastructure, including the policy, which is used to control the connected automated vehicles simulated by SUMO.
The characteristics of the cockpit are summarized in Table \ref{tab:cpit}.
}

\begin{table}[ht]
\begin{center}
\resizebox{0.7\columnwidth}{!}{%
\begin{tabular}{l|l|}
\hline
\textbf{Physical quantity}             & \textbf{Values} \\ \hline
Platform Size                 & 6 meters x 6 meters         \\
Visual System (H)             & 270$^\circ$           \\
Visual System (V)             & 90$^\circ$           \\
Degree of freedom             & 9               \\
Longitudinal Acc. of the Base & 1.5 G-force           \\
Lateral Acc. of the Base      & 1.5 G-force           \\
Vertical Acc. of the Cockpit  & 2.5 G-force           \\
Longitudinal Travel           & 4.2 meters          \\
Lateral Travel                & 4.2 meters           \\
Vertical Travel               & $\pm$ 298 millimeters       \\
Yaw Angle                     & $\pm$ 62$^\circ$        \\
Roll Angle                    & $\pm$ 15$^\circ$         \\
Pitch Angle                   & $\pm$ 15$^\circ$        \\ \hline
\end{tabular}%
}
\caption{A summary of the physical characteristics of the driving simulator (cockpit).}
\label{tab:cpit}
\end{center}
\end{table}

\subsection{Reinforcement Learning}
Reinforcement Learning (RL) constitutes a branch within the domains of Artificial Intelligence and Machine Learning that draws its initial inspiration from the Pavlovian paradigm of conditioning, wherein organisms adapt their behavioral choices based on the gratifications (rewards) and aversions (punishments) received. Analogously, 
in the classic RL setup an artificial agent (sometimes referred to as an actor) is engaged in dynamic interactions with its environment, thereby selecting at discrete time intervals its course of action. RL algorithms aim at training the agent, enabling it to independently attain optimal behaviors within a designated environment, in alignment with contextually defined objectives inherent to the specific problem at hand. The remaining part of this section summarises the fundamental RL definitions (see for a more detailed introduction \cite{sutton2018reinforcement, bertsekas2019reinforcement, bertsekas1996neuro}). The RL problem can be mathematically formalized as a Markov Decision Process, by considering a tuple $(S, \, A, \, p, \, r )$ such that $S$ and $A$ are the sets of possible agent' states and actions, respectively, while $p: S \times A \times S  \longrightarrow [0, \, 1 ]$ indicates the probability to transition from state $s$ to state $s'$ by acting $a$, i.e. $p(s, \, a , \, s' ) = \mathbb{P}( s_{t+1} = s' \vert s_{t} = s, \, a_{t} = a )$. The reward function $r: S \times A  \longrightarrow \mathbb{R}^{+}_{0}$ maps a state-action couple with its immediate, intrinsic desirability in relation to the agent's task at hand. In this context, the agent's decision-making process is modeled as a policy function $\pi: S \times A \longrightarrow [0, \, 1 ]$, such that $\pi(s, \, a)$ represents the probability of acting $a$ while in state $s$, determining \textit{de facto} the agent's behavior. It is possible to define a value function $V_{\pi}: S \longrightarrow \mathbb{R}^{+}_{0}$ associated with any policy $\pi$, that represents the expectation with respect to $p$ of the  cumulative discounted rewards obtained by $\pi$ over time from each state, considering $\gamma  \in ( 0, \, 1 ]$ as discounting factor penalizing rewards obtained further on in time, i.e.,
\begin{equation*}
V_{\pi}(s) = \mathbb{E}_{_{\substack{s_{t+1} \sim p_t,\\ a_t \sim \pi_t}}}\Big[  \sum_{t=0}^{\infty} \gamma^{t} r(s_t, a_t) \, \vert \, s_0 = s \Big], 
\end{equation*}
where we define per each step $t$ the distributions $p_t = p(s_t, a_t, \,\cdot \, )$ and $\pi_t = \pi(s_t)$.
It is sometimes useful to consider the action-value function $Q_{\pi}: S \times A \longrightarrow \mathbb{R}^{+}_{0}$ associated to $\pi$, where $Q_{\pi}(s, \, a)$ indicates the expected cumulative discounted rewards obtained by following policy $\pi$ over an infinite horizon, starting from state $s$ and performing as a first action $a$, i.e.
\begin{equation*}
Q_{\pi}(s, a) = \mathbb{E}_{_{\substack{s_{t+1} \sim p_t,\\ a_t \sim \pi_t }}}\Big[  \, \sum_{t=0}^{\infty} \gamma^{t} r(s_t, a_t) \, \vert \, s_0 = s, a_0 = a  \, \Big],
\end{equation*}
keeping in mind that the following property holds
\begin{equation*}
Q_{\pi}(s, a) = \mathbb{E}_{s' \sim p(s, a, \,\cdot \, )} \Big[ \, r(s, a) + \gamma \, V_{\pi}(s')  \, \Big].
\end{equation*}
The agent's overall goal is to learn a policy $\pi^{*}$ that maximizes the expected long-term desirability of each state, i.e., such that for each $s \in S$
\begin{equation}
\pi^{*}(s) = \text{arg}\max_{\pi} V_{\pi}(s).
\label{eq:RL-pb}
\end{equation}
{\color{black}In many cases it is common to model $r$ as a function linking states and actions with an associated cost (instead of a reward), without loss of generality, and substitute the maximization with a minimization in \eqref{eq:RL-pb}.}

\section{Literature Review}
\label{sec:litrev}
{\color{black}
This section reviews significant contributions in the literature regarding the study of mixed-traffic scenarios and the design of RL policies for AVs. In this work we do not focus on the development of new RL algorithms for AVs training. On the contrary, in order to realistically study the interaction and coexistence of AVs and HVs, we find more sensible to employ established and, hence, more reliable RL techniques, closer to the deployment phase, than extremely recent methods. Hence we introduce the related literature not from the algorithmic point of view, but summarizing some of the most common problem design choices.
}

\subsection{AVs-HVs Interaction}
{\color{black}
Predictions suggest a gradual market integration of AVs, with estimates ranging from 24\% to 87\% market share by 2045 \cite{talebpour2016influence, rahman2018longitudinal}. Most of the studies about the role of AV integration explore how such vehicles will lead to a fundamental transformation in mobility \cite{lee2020regulations,yang2017impact,olia2016assessing,perraki2018evaluation,liu2018modeling}. A number of research papers have delved into the advantages of AVs, and in particular Connected AVs (CAVs), concluding that they can notably decrease traffic incidents, alleviate congestion, lower fuel use, and offer essential mobility options \cite{lee2020regulations,yang2017impact,olia2016assessing,zhang2015exploring,santana2021transitioning, kala2013motion, mahdavian2019assessing, Mavromatis2020, Ye2018Heterogeneous}. Researchers highlighted that even a small penetration of AVs may significantly impact, for example, the number of parking requirements and the number of accidents \cite{zhang2015exploring,yang2017impact}. Other researchers also investigated how street features like  Variable Speed Limits and road capacities impact urban scenarios under different levels of AV integration \cite{perraki2018evaluation, liu2018modeling}. Focusing on the interaction between HVs and AVs, in \cite{schieben2019designing} researchers propose considerations on the design of AVs that can safely and intuitively interact with other traffic participants, based on common human interaction strategies. A driving simulator experiment was conducted in \cite{SCHOENMAKERS2021119} on 34 individuals, to investigate the behavior of HVs exposed to different road design of lanes dedicated to AVs on motorways. The coexistence and interaction of AVs and HVs was investigated as well in \cite{SONI202248}, through a field test conducted on 18 participants, focusing on gap acceptance, car-following, and overtaking behaviors, showing that drivers interacting with recognizable AVs adopt smaller critical gaps, and after overtaking, merge closer in front of those. The work in \cite{REDDY2022451} is devoted to understanding HVs behavior in mixed traffic at un-signaled priority T-intersections, through a driving simulator experiment on 95 human drivers, whose findings suggest that human drivers change their gap acceptance behavior in mixed traffic depending on AVs recognizability and driving style. Finally, the authors of \cite{MAHDINIA2021106006} aimed at quantifying the behavioral changes caused by human drivers following either an AV or an HV, and their impact on safety, fuel consumption and pollution, by analyzing data from a field experiment on 9 drivers. Their work shows that, when human drivers follow AVs, lower driving volatility in terms of speed and acceleration can be achieved, and consequently more stable traffic flow behavior, lower crash risk, less fuel consumption and emission production are experienced. However, to the best of our knowledge, there is no study that evaluates how the penetration rate of AVs may impact the human driver's perception of traffic smoothness, safety and comfort in a scenario in which AVs and HVs coexist. Such evaluation is the main goal of our study.
}

\subsection{RL design for AVs}

{\color{black}}

In the context of RL policies training for AVs, literature can be found detailing the most common choices related to the design of the RL problem to be solved, in particular regarding states, actions and rewards design for AVs training. It is frequent to consider a states space that includes the position, heading, velocity of the vehicle, and the presence of obstacles within the sensor's view, possibly employing a Cartesian or Polar occupancy grid centered around the vehicle. This grid is often enhanced with lane-related details such as lane number, path curvature, historical and predicted trajectory of the vehicle, longitudinal measures such as time-to-collision, and broader scene-related information such as traffic regulations and the locations of traffic signals \cite{leurent2018survey, kiran2020deep}. The action space instead is by definition related to the actuators present on the vehicle and devised to the vehicle control task. Multiple actuators come into play, both continuous (as for instance steering angle, throttle, and brake) and discrete (i.e., gear changes). A framework incorporating temporal abstractions, such as options (sub-policies that extend primitive actions over multiple time steps, to be chosen instead of low-level actions), can simplify action selection \cite{sutton1999between}. {\color{black}Concerning the design of suitable reward functions for RL agents in the context of autonomous driving, researchers follow a variety of approaches, in order to tackle the many different sub-skills that all together characterize the general goal of autonomously driving a vehicle.} Examples include measures such as distance traveled towards a destination \cite{dosovitskiy2017carla}, speed of the vehicle \cite{dosovitskiy2017carla, li2018urban, kardell2017autonomous}, maintaining the ego vehicle at a standstill \cite{chen2019model}, avoiding collisions with other road users or scene objects \cite{dosovitskiy2017carla, li2018urban}, adherence to sidewalk rules \cite{dosovitskiy2017carla}, staying in the lane, ensuring comfort and stability while avoiding extreme acceleration, braking, or steering \cite{kardell2017autonomous, chen2019model}, and adherence to traffic regulations \cite{li2018urban}.

\section{Methodology}
\label{sec:methodology}
This section describes all the steps and experimental setups adopted to carry out the study. First, we describe how SUMO can be turned into a realistic RL environment using Python libraries like Flow \cite{wu2021flow}, Ray RLlib \cite{liang2018rllib} and OpenAI Gym \cite{brockman2016openai}.

\subsection{Turning SUMO into a realistic Reinforcement Learning-based environment}
\label{sec:sim}
In this work, we transform SUMO, a high-fidelity multi-agent transportation simulator, into a realistic Reinforcement Learning environment to optimize and evaluate policies for AVs. The ultimate goal is to use SUMO to learn policies in which AVs take optimal actions to reduce emissions and to minimize the time to cross a real roundabout in Milan, Italy. 

To transform SUMO into a Reinforcement Learning environment, we integrate {\color{black} Flow} \cite{wu2021flow}, an open-source Python package that can be used to create a communication layer between SUMO and Ray RLlib \cite{liang2018rllib}. Remarkably, {\color{black} Flow} can be used to investigate the so-called mixed autonomy scenarios where only a portion of the deployed cars are AVs, and the others are controlled by car following models (CFMs), a set of ordinary differential equations that realistically mimic basic traffic dynamics on single-lane roads. This ability to study mixed autonomy scenarios is paramount for policymakers and traffic engineers as it represents a more realistic short-term scenario. In {\color{black} Flow}, SUMO is connected with Ray RLlib and OpenAI Gym \cite{brockman2016openai} through TraCI, a package to control communication over network protocols. The environment in which agents operate consists of a realistic network representing a physical road layout (e.g., speed limits, lanes, length, shape). The actors are the deployed cars. Some of them (marked with ``rl\_agent'') are controlled by a learned policy and make decisions according to a specific goal they have to minimize. Other cars (marked with ``human\_agent'') base their decision on pre-defined driver models.

\newpage {\color{black} Flow allows to rely on \textit{observer}
functions to map the set of states $S$ to the observations $O$. This permits to tailor the information provided to the controller, choosing a subset of the SUMO states of the vehicle. We structure the observation vector $o_t^n \in O$ available at instant $t$ to vehicle $n$ so that it contains the last measured value of position $x_t^n$ and acceleration $\ddot{x}_t^n$ of the vehicle. Moreover, we include in $o_t^n$ information obtained from the estimated position and acceleration of its front ($F$), back ($B$), left ($L$) and right ($R$) neighbors in the scenario, indicated respectively as $\lbrace x_t^k, \ddot{x}_t^k \rbrace_k$ with $k = F, B, L, R$, if such vehicles are present at instant $t$ around vehicle $n$. The values associated with the neighbors that are missing at time $t$ are replaced by an opportune placeholder. The described components of the observations vector are divided by quantities that are characteristic of the chosen scenario, such as the scenario's dimension $x_{\text{max}}$ and the maximum acceleration $\ddot{x}_{\text{max}}$. Summarizing, we consider as observation $o_t^n$ the vector
\begin{equation}
    o_t^n = 
    \begin{bmatrix}
        x_t^n / x_{\text{max}} \\
        \ddot{x}_t^n / \ddot{x}_{\text{max}} \\
        x_t^F / x_{\text{max}} \\
        (\ddot{x}_t^F - \ddot{x}_t^n) / \ddot{x}_{\text{max}} \\
        x_t^B / x_{\text{max}} \\
        (\ddot{x}_t^B - \ddot{x}_t^n) / \ddot{x}_{\text{max}} \\
        x_t^L / x_{\text{max}} \\
        (\ddot{x}_t^L - \ddot{x}_t^n) / \ddot{x}_{\text{max}} \\
        x_t^R / x_{\text{max}} \\
        (\ddot{x}_t^R - \ddot{x}_t^n) / \ddot{x}_{\text{max}}
    \end{bmatrix}.
\end{equation}
The action $a_t^n \in A$ decided by agent $n$ at time $t$, instead, consists of the next acceleration value $\ddot{x}_{t+1}^n$ to actuate, and of a discrete decision $c_{t+1}^n \in \lbrace 0, 1 \rbrace$, corresponding to changing the line/maintaining the current one, that is
\begin{equation}
a_t^n = 
    \begin{bmatrix}
       \ddot{x}_{t+1}^n \\
       c_{t+1}^n
    \end{bmatrix}
\end{equation}
}
Our work aims at leveraging the presented infrastructure to allow AVs to learn {\color{black} via RL techniques} the optimal control law to cross a roundabout as fast as possible. In this sense, we design {\color{black} a stage-cost function $r: \mathcal{O} \times \mathcal{A}
\rightarrow \mathbb{R}_{\geq 0}$ measuring the deviation $d$ of the AVs velocity $\dot{x}_t^n$ from a user-defined desired velocity $v$. In order to penalize the early termination of roll-outs due to collisions or other failures, we subtract such deviation from the peak allowable deviation $d_{\text{max}}(v)$. Finally, to ensure non-negativity, the cost is then bounded below by $0$, i.e., 
\begin{equation}
    r_v(\, o_t^n, \, a_t^n \, ) = \max(\, 0, \, \vert\vert \, d(\,\dot{x}_t^n, \, v \,) \, - \, d_{\text{max}}(v) \, \vert\vert \, ).
\end{equation}
Simultaneously, we are interested in reducing polluting emissions. Hence, we leverage the previously described velocity-based stage-cost with an analogous one, punishing the deviation from a target level of pollution $P$, i.e.,  $d(p_t^n, P)$,
where the pollution levels $p_t^n$ are measured in function of the actuated decisions $a_t^n$, and $P$ is estimated by considering a single vehicle in the scenario, running at constant velocity. The two components of the cost are summed and normalized, and then assigned as a stage-cost to the AV.} We use the well-known Proximal Policy Optimization \cite{shulman2017proximal} algorithm (see Section \ref{sec:ppo} for more details) to learn a policy allowing the AVs to decide at each instant which action to take, based on the agent's individual observations' vector {\color{black} $o_t^n \in O$. All the AVs involved in the roundabout scenario collaborate in training a central policy by providing their simulated experience in order to update the policy parameters. Moreover, during training and deployment, when interrogated, the policy exploits both the local observations of the agent taking the decision and information related to the position and \color{black}acceleration of other vehicles in the roundabout that are received through a central communication scheme.} 

Differently from the state-of-the-art studies, we evaluate the learned policy quantitatively and qualitatively{, analyzing results obtained through} tests carried out at the DriSMi Laboratory of  Polytechnic University of
Milan. This laboratory has a high-fidelity last-generation cable-driven driving simulator (see Figure \ref{fig:simulator} and \cite{previati2023roundabout, previati2024cooperative}). The traffic scenario is generated thanks to a co-simulation between SUMO (traffic engine), and VI-WorldSim\footnote{\href{https://www.vi-grade.com/en/products/vi-worldsim/}{https://www.vi-grade.com/en/products/vi-worldsim/}}, that simulates the motion of the vehicle driven by the human in the loop and allows to have a graphical representation of the traffic scenario.  The combination of SUMO and VI-WorldSim allows us to evaluate the learned policy in terms of the traffic smoothness and safety perception of the passengers (i.e., real humans who agreed to participate in the experiments in our case). Further details about the evaluation procedure are shared in Sections \ref{sec:qt_eval} and \ref{sec:ql_eval}.

\subsection{Proximal Policy Optimization}
\label{sec:ppo}
To learn the policy parameters, we employ the Proximal Policy Optimization (PPO) method \cite{shulman2017proximal}. PPO is a prominent policy-optimization algorithm, deriving from the Trust Region Policy Optimization (TRPO) algorithm \cite{schulman2015trust}, and improving it from the flexibility and computational complexity point of view. Both PPO and TRPO are designed to stably and efficiently optimize decision policies, focusing on refining policies by iteratively adjusting their parameters while limiting the extent of these updates in order to maintain stability during learning. This is realized by PPO within a dual-step process: policy evaluation and policy improvement. During policy evaluation, data are gathered by executing the current policy within the environment. Subsequently, the advantages 
$$A_{\pi_{\theta}}(s, a) = Q_{\pi_{\theta}}(s, a) - V_{\pi_{\theta}}(s)$$
of actions taken with respect to expected returns are computed. The advantages serve as a measure of how favorable the chosen actions were, with respect to expected outcomes. 
Following policy evaluation, policy improvement is performed through several epochs of optimization. In each epoch, PPO computes surrogate objectives that quantify the change in the policy's performance with respect to the previously considered set of policy parameters $\theta_{\text{old}}$, guided by the advantage values, i.e.,
\begin{equation*}
L_{\theta}(s, a) = \min\left(\,\,  
\frac{\pi_{\theta}(a|s)}{\pi_{\theta_{\text{old}}}(a|s)} \, A_{\pi_{\theta}}(s, a), \,\,\,\, \text{C} \, A_{\pi_{\theta}}(s, a) \,\, \right), 
\end{equation*}
with 
\begin{equation*}
    \text{C} = \text{clip}\left(\frac{\pi_{\theta}(a|s)}{\pi_{\theta_{\text{old}}}(a|s)},   \,\, 1 - \epsilon,  \, \, 1 + \epsilon\right).
\end{equation*} 
These surrogate objectives facilitate the optimization of the policy in a manner that promotes positive action shifts while maintaining a threshold $\epsilon$ on the magnitude of policy updates. This threshold, referred to as the ``clip parameter,'' curbs policy updates from straying too far from the original distribution, ensuring a measure of stability and preventing drastic policy shifts that could lead to instability. PPO's distinctive feature lies in its capacity to strike a balance between exploiting the advantages of updated policies and maintaining a controlled adjustment process. By constraining policy updates and employing the surrogate objective, PPO achieves stable and incremental policy improvements, contributing to efficient and reliable RL.

{\color{black} These characteristics, together with the performance attained by PPO in many benchmark examples and applications \cite{yu2022surprising}, motivated us to choose PPO as the learning algorithm for AVs policy optimization. Moreover, in order to conduct an objective evaluation of the effects of deploying AVs on realistic scenarios in the presence of human drivers, technologies that are both well-performing and well established, as for instance PPO, should be preferred to algorithms that, although cutting-edge, are not at maturity, and hence are less commonly deployed in real applications.} 

\subsection{Roundabout Design and Data Collection}
\begin{figure}
    \centering
fpollu    \includegraphics[width=1\linewidth]{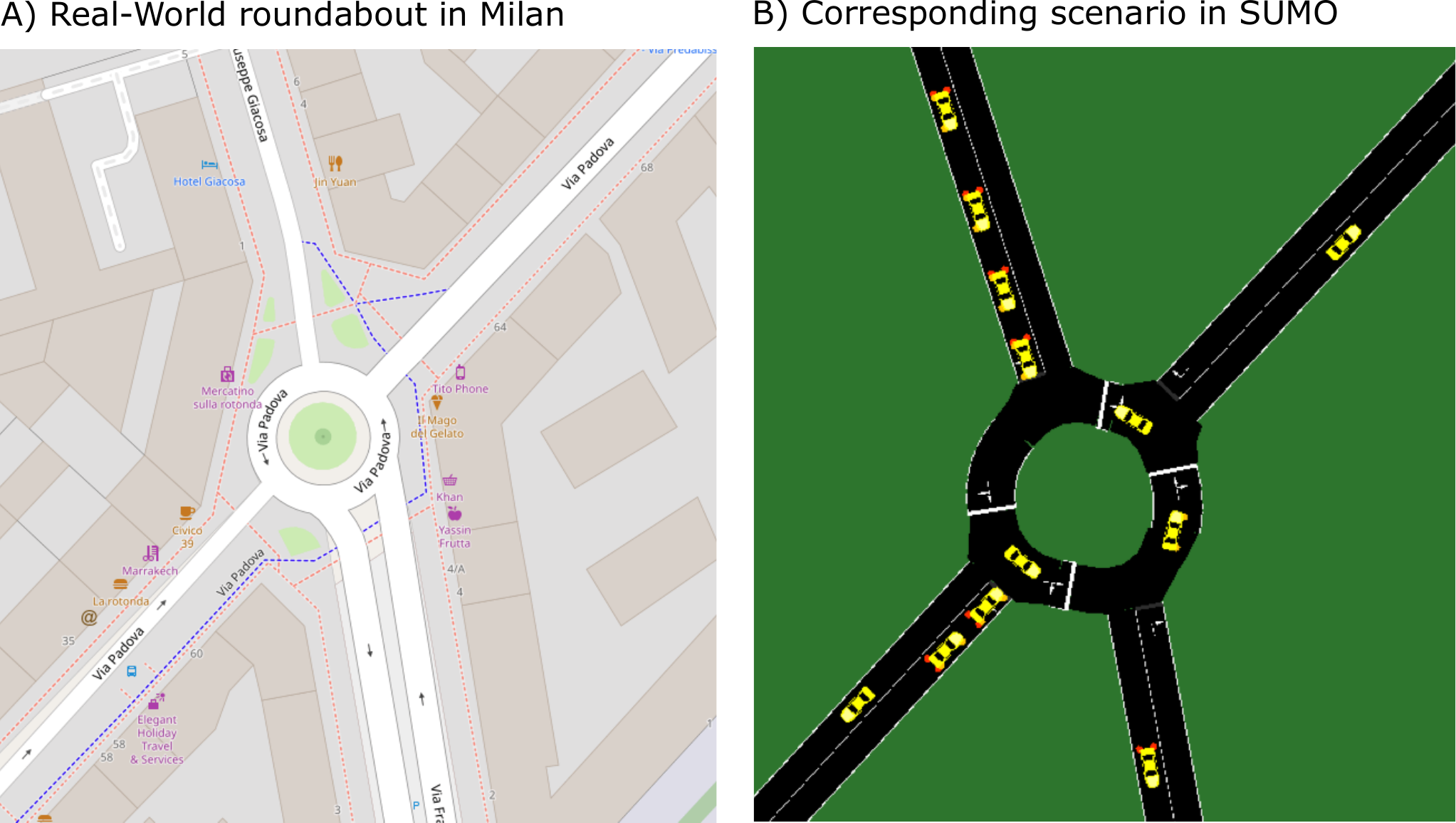}
    \caption{\textbf{A)} The selected roundabout in Milan observed from OpenStreetMap; and in \textbf{B)}, the corresponding design of the roundabout in SUMO. Also, realistic traffic loads for fine-tuning have been measured through field measurements.}
    \label{fig:roundabout}
\end{figure}

To operate in an environment that is as realistic as possible, we designed a roundabout inspired by a real-world one in Milan, Italy. In Figure \ref{fig:roundabout} A), we can see how the roundabout looks in the real world, while in panel B), we can see the SUMO's roundabout. It is a four-leg mini-roundabout, showing medium-high traffic and, therefore, being a challenging environment for the AV policy. Moreover, it has some important details, which make this particular scenario of general interest, specifically:
\begin{itemize}
    \item every leg has pedestrian crosswalks immediately before the entrance of vehicles inside the circulatory roadway.
    \item  Two of the legs are central arteries of the city, greatly increasing traffic on the roundabout.
    \item The roundabout has a standard configuration widely distributed in European urban areas \cite{inbook} with significant flows.
\end{itemize}

A calibration procedure was conducted to replicate the number of vehicles approaching the intersection and their positions during the simulation. Firstly, measurements were taken for the maximum queue length, upstream and downstream flows for each leg, considering road vehicles, pedestrians, and bicycles on the actual roundabout. This process was repeated for six consecutive time slots, each lasting 10 minutes.

Subsequently, the results of these measurements were compared with simulations conducted in SUMO to calibrate the seven most relevant parameters that define the traffic conditions in the considered scenario. The parameters under consideration include the distance at which a pedestrian is considered, the minimum time interval to cross the path of another vehicle when entering the roundabout, the time before a driver enters the roundabout even if obstructing the way of an incoming vehicle, the maximum acceleration and deceleration of the vehicles, the time interval between vehicles, and the drivers’ reaction time. For each simulation run, a cost function was constructed to compare the mean and maximum queue lengths between the measured and simulated data. The calibrated parameters are those that minimize the differences between the two sets of data. 
{
Specifically, Table \ref{tab:sumo_params} lists the SUMO parameters detailed below and their values: 
\begin{itemize}
    \item \textbf{jmCrossingGap}: minimum distance between the vehicle and the pedestrian that is heading toward the point of conflict of its trajectory with that of the vehicle;
    \item \textbf{jmTimegapMinor}: minimum time interval for a vehicle to enter an intersection where it does not have the right-of-way, before a vehicle with right-of-way;
    \item \textbf{impatience}: driver’s intent to obstruct a vehicle with the right of way;
    \item \textbf{accel}: maximum acceleration for the selected vehicle type;
    \item \textbf{decel}: maximum deceleration for the selected vehicle type;
    \item \textbf{tau}: minimum time interval between consecutive vehicles;
    \item \textbf{actionStepLength}: driver reaction time.
\end{itemize}

\begin{table}[]
\center
\resizebox{0.5\columnwidth}{!}{%
\begin{tabular}{l|l|}
\hline
\multicolumn{1}{c|}{{Parameter}} & {Value} \\ \hline
jmCrossingGap                           & 1,3545         \\
jmTimegapMinor                          & 1,7792         \\
impatience                              & 0,1182         \\
accel                                   & 1,7634         \\
decel                                   & 4,2939         \\
tau                                     & 1,3472         \\
actionStepLength                        & 0,505          \\ \hline
\end{tabular}%
}
\caption{SUMO calibrated parameters}
\label{tab:sumo_params}
\end{table}

}

As can be noted from \ref{fig:roundabout} B), although present in the real roundabout, in the final network there are no restricted lanes or pedestrian crosswalks. Such elements have been removed, after the calibration process, as they are not within the scope of the AI@EDGE project. Correction coefficients, taken from the literature, have been used to take these elements into account in the modified model. For the same reason, only cars are considered; other vehicles and pedestrians are considered via equivalent coefficients \cite{giuffre2018capacity}.

\section{Results and Discussion}
\label{sec:res}

\begin{table}[ht!]
\begin{center}
\resizebox{0.8\columnwidth}{!}{
\begin{tabular}{l|ll|ll|}
\hline
\%AVs & $\mu$ AVs ($s$) & $\mu$ HVs ($s$) & \#AVs  & \#HVs  \\ \hline
0      & -           & 17.94       & 0    & 1540 \\
10     & 17.15       & 17.32       & 154  & 1386 \\
20     & 17.02       & 17.17       & 308  & 1232 \\
30     & 16.92       & 17.12       & 462  & 1078 \\
40     & 16.76       & 17.08       & 616  & 924  \\
50     & 16.58       & 16.99       & 770  & 770  \\
60     & 16.33       & 16.75       & 924  & 616  \\
70     & 16.19       & 16.66       & 1078 & 462  \\
80     & 15.98       & 16.63       & 1232 & 308  \\
90     & 15.74       & 16.41       & 1386 & 154  \\
100    & 15.31       & -           & 1540 & 0    \\ \hline
\end{tabular}
}
\caption{\label{tab:time_cross} Given a certain penetration rate of AVs (column \%AVs), we measure the average time that AVs and HVs need to cross the roundabout as in Eq.\eqref{crossing-time-eq}. The measurements are in seconds. We observe that as the number of AVs (column \#AVs) increases, both AVs and HVs, on average, need less time to cross the roundabout, highlighting how HVs may also benefit from the optimization and diffusion of AVs.}
\end{center}
\end{table}

\subsection{Quantitative Evaluation}
\label{sec:qt_eval}
\begin{figure*}
    \centering
    \includegraphics[width=1\linewidth]{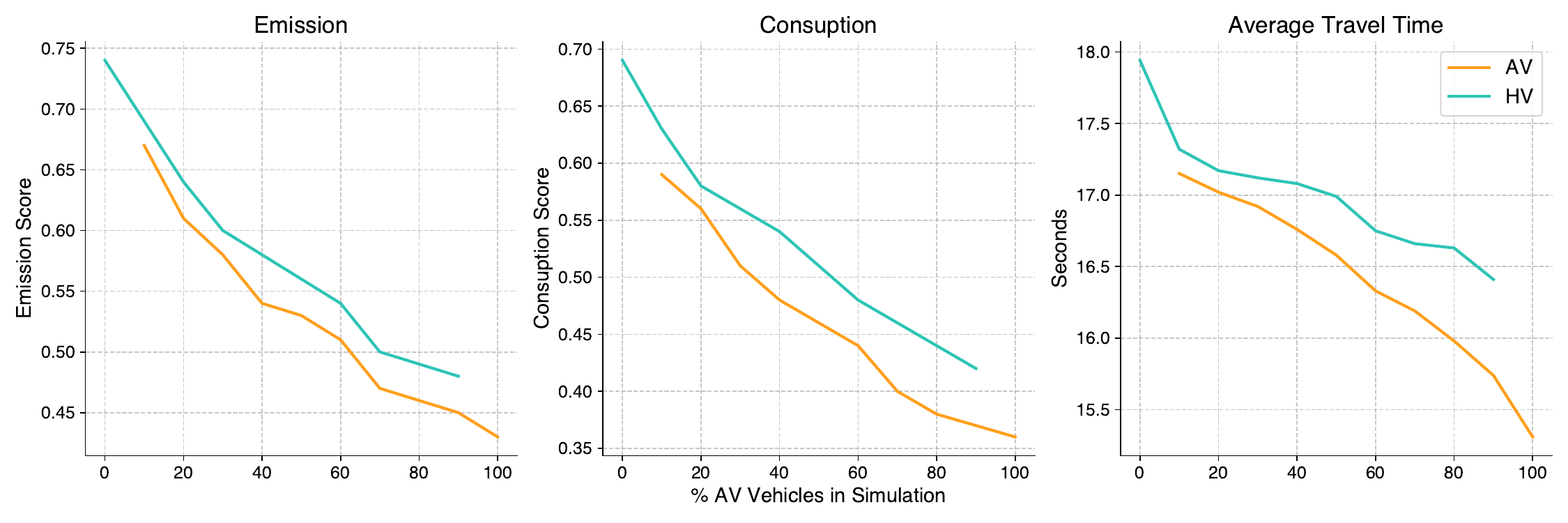}
    \caption{ Concerning the quantitative analysis, we measured the impact of AV penetration rate on emissions, fuel consumption and average time to cross the scenario. \textbf{On the left}, we have the results related to emissions. \textbf{In the middle}, we show the results for fuel consumption while \textbf{on the right} we have the average crossing time. As we can observe, the best results are obtained for the scenario with 100\% of AVs. However, having some AVs (e.g., 10\% to 90\% penetration rate already provides some benefits. Interestingly enough, also HV benefits from the penetration rate of AVs.} 
    \label{fig:quant_eval}
\end{figure*}

In our work, we focus on evaluating the efficacy of the learned policy in reducing the time AVs need to cross the roundabout and emission and fuel consumption. Concerning the former, given a simulation $\mathcal{S}$ of 3600 seconds (i.e., 1 hour) with a given percentage of AVs and a total of $n$ vehicles $v_1, v_2, \dots, v_n$, we associate to each vehicle $v_i$ its entering time $t_{v_i}^{\text{in}}$ and exiting time $t_{v_i}^{\text{out}}$. The average time needed by the cars in the simulation $\mathcal{S}$ to cross the roundabout is then computed as 
\begin{equation}
    \mu(\mathcal{S}) = \frac{1}{n}\,\,\sum_{i=1}^{n} ( \, t_{v_i}^{\text{out}} - t_{v_i}^{\text{in}} \, ).
    \label{crossing-time-eq}
\end{equation}
Note that by sampling $V=\lbrace v_1, v_2, \dots, v_n \rbrace$ it is possible to estimate the average time for AVs and HVs. As a reminder, the simulation is carried out to have a total of 1540 vehicles passing through the roundabout over one hour as emerged from the field measurements. Time is measured in seconds.
In Table \ref{tab:time_cross} and {in the right panel of Figure \ref{fig:quant_eval}}, we can see the results when having a different amount of AVs leveraging the learned policy.

At 0\% AV penetration, only HVs are present. The average time taken by HVs to cross the roundabout stands at 17.94 seconds.
As we introduce AVs into the system, there's a slight improvement in the average crossing times for both types of vehicles. At 10\% AV penetration, the average crossing time for AVs is 17.15 seconds. The HVs also experience a marginal decrease in their average time, clocking in at 17.32 seconds. As the percentage of AVs increases from 10\% to 50\%, there is a gradual and consistent decrease in the average crossing times for both AVs and HVs. At the midpoint, with 50\% AVs and 50\% HVs, AVs take an average of 16.58 seconds, while HVs take a slightly longer 16.99 seconds.
 Beyond the 50\% mark, as AVs become more dominant, their average crossing times continue to decrease, reaching 15.31 seconds at 100\% penetration. Interestingly, the HVs also benefit from the increased AV presence. At 90\% AV penetration, when only a small fraction of HVs remain, their average crossing time reduces to 16.41 seconds - the minimum reached value.
Throughout the progression, AVs consistently demonstrate a trend of reduced average crossing times with their increased presence. HVs also benefit from the introduction of AVs. The data suggests that as the percentage of AVs in traffic increases, the roundabout navigation becomes more efficient for both vehicle types. This could be attributed to the predictability and coordination of AVs, which seems to not only benefit their kind but also aid in optimizing the flow for human drivers. 

Another evaluation metric that the policy should optimize is fuel consumption and pollution. Both information can be extracted from each vehicle's property and computed by SUMO. To give an idea of how the model is performing, instead of using the values provided by SUMO we report a more interpretable score normalized between $0$ and $1$. To this end, for each scenario $\mathcal{S}$, we take the worst-performing vehicle ($\text{score}_{\text{max}}$) and the best-performing vehicle ($\text{score}_{\text{min}}$) as normalizing factors. We then performed a min-max normalization, ending up with a score between $0$ and $1$ for each vehicle, with $0$ indicating lower emission or lower fuel consumption and $1$ indicating the worst performance. { In particular, the adopted formula is: 
\begin{equation}
    \mu(\mathcal{S}) = \frac{1}{n}\,\, \sum_{i=1}^{n}\Big( \frac{\text{score}(v_i)-\text{score}_{\text{min}}}
    {\text{score}_{\text{max}}-\text{score}_{\text{min}}}\Big).
    \label{normalized-score-eq}
\end{equation}
where $n$ is the number of vehicles in simulation $\mathcal{S}$ and the term $\text{score}(v_i)$ in the formula can indicate both the emission or the consumption level of the $i$-th vehicle $v_i$.
}
The results can be seen in Table \ref{tab:fuel_score} {and in the left -emissions- and center -consuption- panel of Figure \ref{fig:quant_eval}.}

\begin{table}[ht!]
\resizebox{\columnwidth}{!}{%
\begin{tabular}{l|ll|ll|ll|}
\hline
\multicolumn{1}{c|}{\multirow{2}{*}{\%AVs}} & \multicolumn{2}{c|}{AV}                                        & \multicolumn{2}{c|}{HV}                                        & \multicolumn{1}{c}{}    & \multicolumn{1}{c|}{}    \\
\multicolumn{1}{c|}{}                        & \multicolumn{1}{c}{consumption} & \multicolumn{1}{c|}{emission} & \multicolumn{1}{c}{consumption} & \multicolumn{1}{c|}{emission} & \multicolumn{1}{c}{\#AVs} & \multicolumn{1}{c|}{\#HVs} \\ \hline
0                                            & -                              & -                             & 0.74                           & 0.69                          & 0                       & 1540                     \\
10                                           & 0.67                           & 0.59                          & 0.69                           & 0.63                          & 154                     & 1386                     \\
20                                           & 0.61                           & 0.56                          & 0.64                           & 0.58                          & 308                     & 1232                     \\
30                                           & 0.58                           & 0.51                          & 0.60                           & 0.56                          & 462                     & 1078                     \\
40                                           & 0.54                           & 0.48                          & 0.58                           & 0.54                          & 616                     & 924                      \\
50                                           & 0.53                           & 0.46                          & 0.56                           & 0.51                          & 770                     & 770                      \\
60                                           & 0.51                           & 0.44                          & 0.54                           & 0.48                          & 924                     & 616                      \\
70                                           & 0.47                           & 0.40                          & 0.50                           & 0.46                          & 1078                    & 462                      \\
80                                           & 0.46                           & 0.38                          & 0.49                           & 0.44                          & 1232                    & 308                      \\
90                                           & 0.45                           & 0.37                          & 0.48                           & 0.42                          & 1386                    & 154                      \\
100                                          & 0.43                           & 0.36                          & -                              & -                             & 1540                    & 0                        \\ \hline
\end{tabular}%
}
\caption{\label{tab:fuel_score} Given a certain penetration rate of AVs (column \%AVs), we measure the consumption score and emission score of both AVs and HVs, following Eq.\ref{normalized-score-eq}. Similarly to what happened with the crossing time, we observe that as the number of AVs (column \#AVs) increases, both AVs and HVs, on average, consume less fuel and reduce their emissions.}
\end{table}

When no AVs are present (0\% AV penetration), the average fuel consumption score for HVs stands at 0.74, with an emissions score at 0.69.
 As AVs are introduced into the system at 10\% penetration, their fuel consumption and emissions are recorded at 0.67 and 0.59, respectively. Interestingly, even with a modest AV penetration, there's an observable improvement in HV metrics as well. The consumption for HVs drops to 0.69, and their emissions decrease to 0.63. As the percentage of AVs in the system grows from 10\% to 50\%, there's a consistent improvement in both consumption and emissions for both vehicle types. At 50\% AV penetration, AVs record consumption and emission values of 0.53 and 0.46, respectively, while HVs show values of 0.56 and 0.51. Beyond 50\% AV penetration, the trend of decreasing consumption and emissions continues for AVs, reaching their lowest at 0.43 and 0.36, respectively at 100\% AV penetration. HVs also see continuous improvement. Throughout the entire range, AVs exhibit lower consumption and emission values compared to HVs. The difference becomes more pronounced as the percentage of AVs increases, highlighting the efficiency of autonomous driving systems. In conclusion, not only do AVs demonstrate better fuel efficiency and lower emissions, but their presence also seems to positively influence the performance of human-driven vehicles, similar to travel time reduction. This could be attributed to smoother and more predictable traffic flow, leading to less stop-and-go traffic and more consistent driving speeds learned by the policy.

\subsection{Qualitative Evaluation}
\label{sec:ql_eval}
{ As previously mentioned, }tests have been conducted at the DriSMi Laboratory of the Polytechnic University of
Milan\footnote{\href{https://www.drismi.polimi.it/drismi/}{https://www.drismi.polimi.it/drismi/}} {to qualitatively evaluate indicators of the passengers' comfort in the described driving scenario, considering different AV penetration levels}. The laboratory has a high-fidelity, last-generation driving simulator as described in Section \ref{sec:cpit}. The integration between SUMO, VI-WorldSim and the simulator itself allows us to evaluate, to the best of our knowledge for the first time, traffic smoothness and safety perception of passengers as qualitative metrics to analyze the passengers' comfort. To collect the necessary information for our study, a panel of ten participants has been selected for the preliminary tests. The participants were chosen from individuals without previous experience with driving simulators. The panel consists of 5 females and 5 males, aged between 22 and 33 years, with driving experience ranging from 1 to 15 years. Before the test, each participant was given instructions on how to operate the driving simulator and signed an informed consent form. Additionally, each participant spent about ten minutes driving in a simple motorway scenario to become familiar with the driving simulator before the actual test.
Individuals were exposed in simulation to experiences in two driving scenarios, characterized by the  20\% and 80\% of AV penetration rate, respectively. Following the simulated ride, the participants were asked to fill out a quick survey, consisting of three close-ended questions:
\begin{itemize}
    \item Regarding traffic smoothness, which of the following statements do you agree with the most?
    \item Regarding safety perception, which of the following statements do you agree with the most?
    \item Globally, which of the two scenarios did you prefer?
\end{itemize}
{ Per each question, the interviewed individuals were asked to select one among five possible alternatives,} summarized in the following three tables {(Table \ref{tab:q1}, \ref{tab:q2} and \ref{tab:q3} associated with the first, second, and third question, respectively)}. 
\begin{table}[ht!]
\resizebox{\columnwidth}{!}{%
\begin{tabular}{l|l}
\hline
\textbf{Regarding traffic smoothness, which of the}  \\ \textbf{following statements do you agree} \\ \textbf{with the most? }         & \textbf{Number of answers} \\
\hline
Traffic in the scenario with 20\% of AVs was \\ definitely smoother than in the scenario with \\  80\% AVs   & 1                 \\ 
Traffic in the scenario with 20\% of AVs was \\ partially smoother than in the scenario with \\ 80\% AVs     & 3                 \\
Traffic in the scenario with 20\% of AVs was \\ partially less smooth than in the scenario \\ with 80\% AVs  & 4                 \\
Traffic in the scenario with 20\% of AVs was \\ definitely less smooth than in the scenario \\ with 80\% AVs & 1                 \\
I did not perceive differences                                                                         & 1                \\ 
\hline
\end{tabular}%
}

\caption{\label{tab:q1} A summary of the answers to the first question of the survey. As we can see, most of the voters say that the scenario with 80\% of AVs was smoother with respect to the one with 20\% of AVs. One participant did not perceive differences between the two scenarios and 40\% of participants preferred the scenario with 20\% of AVs, although the majority of them only partially.}
\end{table}
In total we collect feedback from 10 individuals. While the sample size is limited, a few tentative conclusions can be drawn. In Table \ref{tab:q1}, most of the respondents (5 out of 10) felt that the traffic was smoother in the scenario with 80\% AVs, to varying degrees. A smaller segment (4 out of 10) felt the opposite, indicating that the 20\% AV scenario was smoother. Just 1 out of 10 respondents did not perceive any significant difference in traffic smoothness between the two scenarios. Results suggest that as AV penetration increases, the perceived smoothness of traffic might improve. However, divergent opinions underscore the complexity of human perceptions and the subjective nature of such assessments.
\begin{table}[ht!]
\resizebox{\columnwidth}{!}{%
\begin{tabular}{l|l}
\hline
\textbf{Regarding safety perception, which of the}  \\ \textbf{following statements do you agree} \\ \textbf{with the most? }         & \textbf{Number of answers} \\
\hline
Traffic in the scenario with 20\% of AVs was \\ definitely safer than in the scenario with \\  80\% AVs   & 0                 \\ 
Traffic in the scenario with 20\% of AVs was \\ partially safer than in the scenario with \\ 80\% AVs     & 2                 \\
Traffic in the scenario with 20\% of AVs was \\ partially less safe than in the scenario \\ with 80\% AVs  & 2                 \\
Traffic in the scenario with 20\% of AVs was \\ definitely less safe than in the scenario \\ with 80\% AVs & 6                 \\
I did not perceive differences                                                                         & 0                \\ 
\hline
\end{tabular}%
}
\caption{\label{tab:q2} A summary of the answers to the second question of the survey. According to 80\% of the participants, the scenario with 80\% of AVs is perceived safer than the scenario with 20\% of AVs. Zero participants did not perceive differences between the two scenarios and 20\% of participants partially preferred the scenario with 20\% of AVs.}
\end{table}

Table \ref{tab:q2} shows that a combined total of 8 respondents felt that the scenario with 80\% AVs was safer to some degree compared to the 20\% AV scenario. On the contrary, only 2 out of 10 respondents felt that the 20\% AV scenario was safer in any capacity. All the participants perceived a difference in safety between the two scenarios. These results suggest an inclination towards perceiving higher AV penetration as safer. { Analogously to what was observed while commenting the perception of smoothness in traffic, even when it comes to safety} the diversity in opinions highlights that safety perceptions can be subjective and can vary among individuals. 

\begin{table}[ht!]
\resizebox{\columnwidth}{!}{%
\begin{tabular}{l|l}
\hline
\textbf{Globally, which of the two scenarios}  \\ \textbf{did you prefer?}          & \textbf{Number of answers} \\
\hline
I definitely preferred the scenario with 20\% AVs \\ with respect to the scenario with 80\% AVs   & 0                 \\ 
I partially preferred the scenario with 20\% AVs \\ with respect to the scenario with 80\% AVs    & 3                 \\
I partially preferred the scenario with 80\% AVs  \\ with respect to the scenario with 20\% AVs  & 3                 \\
I definitely preferred the scenario with 80\% AVs  \\ with respect to the scenario with 20\% AVs & 4                 \\
I cannot say which scenario I preferred                                                                        & 0                \\ 
\hline
\end{tabular}%
}
\caption{\label{tab:q3} A summary of the answers to the third question of the survey. From a general perspective, 70\% of participants preferred the scenario with 80\% of AVs while 30\% preferred the scenario with 20\% of AVs, although only partially.}
\end{table}

Finally, Table \ref{tab:q3} shows that, overall, these results hint at a leaning towards the 80\% AV scenario (as shown by the 7 over 10 individuals preferring it either definitely or partially, versus only 3 just partially preferring the 20\% AV scenario), suggesting that most participants might perceive benefits in scenarios with higher AV penetration. Naturally, { even in this case} the variety of responses also emphasizes the subjective nature of such preferences, and individual perceptions can vary based on personal experiences or beliefs. Further research could delve deeper into the reasons behind these preferences, providing a more comprehensive understanding of public sentiment toward AV integration{, possibly by benefiting of a larger sample of collected feedback from individuals}.

\section{Conclusion}
\label{sec:conclusion}
{\color{black}This research has brought to light significant insights into how human drivers may perceive in terms of safety and comfort the integration of AVs in complex urban traffic scenarios, specifically examining a roundabout in Milan, Italy. Our approach leveraged state-of-the-art traffic simulation tools like SUMO, VI-WorldSim, well-established RL methods and a high-fidelity cockpit to understand the dynamics of AVs penetrations alongside HVs.

The findings of our study underscore the substantial benefits that AVs offer in mitigating common urban challenges such as pollution and traffic congestion. By employing established RL techniques, notably the PPO algorithm, we were able to model and analyze the behavior of AVs in realistic traffic settings. The simulation results are promising, showing that AVs enhance their operational efficiency and positively influence the overall traffic flow, benefiting HVs in the process.

Our primary objective included conducting qualitative assessments with human participants. These assessments revealed a notable preference for environments where autonomous vehicles (AVs) are more prevalent, attributing this preference to enhanced safety and smoother traffic flow. Conducting such evaluations is crucial, as AVs and human-operated vehicles (HVs) will increasingly share roads in the future. Notably, previous research has not explored human drivers' perceptions of AVs and HVs coexisting in a realistic, comprehensive manner, as our proposed framework does. Our findings are in line with quantitative data, indicating a future where AVs and HVs can coexist more seamlessly, leading to safer and more efficient urban environments.

While our study has made strides in understanding the potential of AVs in urban settings, it also opens avenues for future research. Further exploration into alternative reinforcement learning algorithms could provide deeper insights into the optimization of AV behavior. Additionally, expanding the scope of human-in-the-loop evaluations with a larger and more diverse participant pool would be invaluable in enriching our understanding of public perception and acceptance of AVs. Such studies could also explore other aspects of urban mobility impacted by AV integration, including pedestrian safety and public transportation systems.
}
\section*{Acknowledgments}
This work has been partially funded by EU H2020 Research and  Innovation programme  AI@EDGE  under  Grant  Agreement  No  101015922  (https://aiatedge.eu/).
G.S. and B.L. acknowledge the support of the PNRR ICSC National Research Centre for High-Performance Computing, Big Data, and Quantum Computing (CN00000013), under the NRRP MUR program funded by the NextGenerationEU.
\bibliographystyle{IEEEtran}
\bibliography{IEEEabrv, mybib}

\begin{IEEEbiography}[{\includegraphics[width=1in,height=1.25in,trim=0 400 400 400, clip,keepaspectratio]{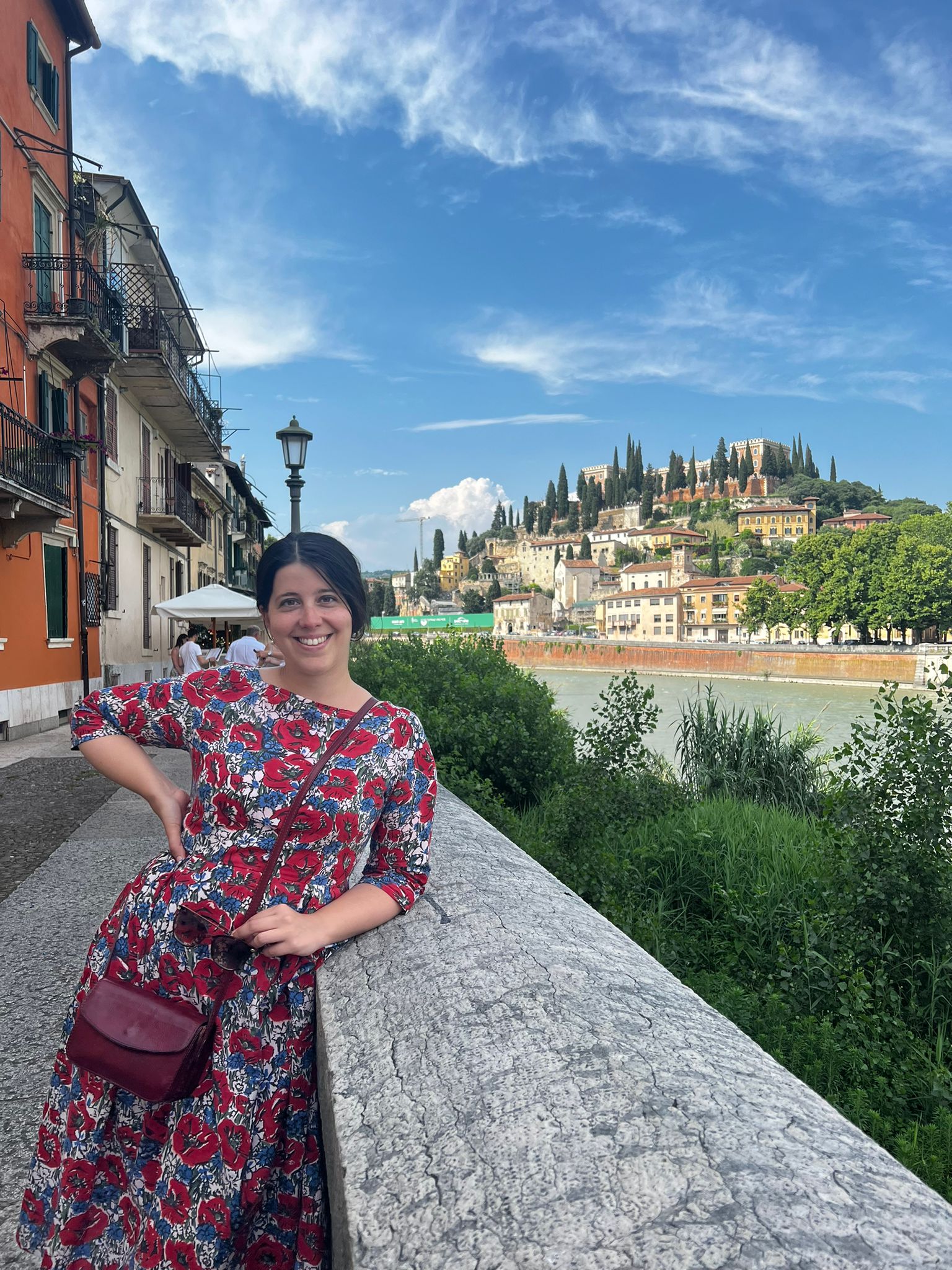}}]{Laura Ferrarotti} is a post-doctoral researcher in the Mobile and Social Computing Lab at Fondazione Bruno Kessler. She received a bachelor’s degree in Mathematics from University of Pisa, a master’s degree cum laude in Mathematical Engineering from Politecnico di Torino, and a Ph.D. in Computer Science and System Engineering at IMT School for Advanced Studies, Lucca, where she worked in the Dynamical Systems, Control and Optimization research unit, on algorithms' design for the synthesis of data-driven controllers based on Reinforcement Learning, and their adaptation to a multi-agent collaborative learning scenario. Her main research interests are Cooperative AI, Federated Learning, Reinforcement Learning, data-driven control, and socially-inspired multi-agent Reinforcement Learning.
\end{IEEEbiography}

\begin{IEEEbiography}[{\includegraphics[width=1in,height=1.25in,clip,keepaspectratio]{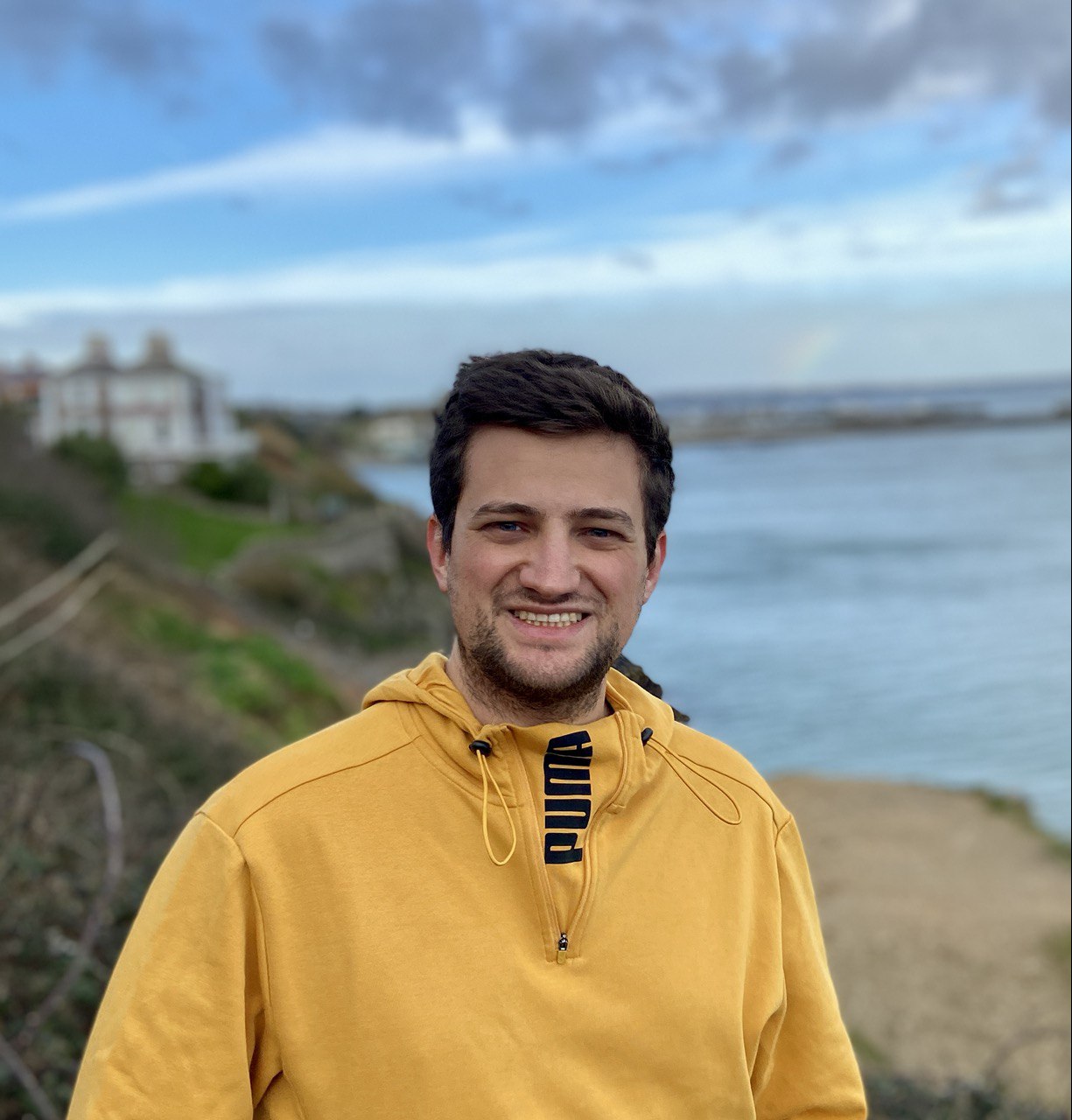}}]{Massimiliano Luca} is a Ph.D. candidate in the Faculty of Computer Science at the Free University of Bolzano and in the Mobile and Social Computing Lab at Fondazione Bruno Kessler. During his Ph.D., Massimiliano had the opportunity to visit and collaborate with institutions like the Massachusetts Institute of Technology (MIT), Centro Nazionale delle Ricerche (CNR) and the Institute for Cross-Disciplinary Physics and Complex Systems - Consejo Superior de Investigaciones Científicas (IFISC / CSIC). Also, Massimiliano is affiliated with the Social Data Science Hub at the University of Edinburgh.
He is broadly interested in the generalization and geographical transferability capabilities of spatio-temporal deep learning models. Previously, he was a research intern at Centro Ricerche Fiat (now Stellantis) and Telefonica Research.
\end{IEEEbiography}
\vskip -2\baselineskip plus -1fil
\begin{IEEEbiography}
[{\includegraphics[width=1in,height=1.25in,clip,keepaspectratio]{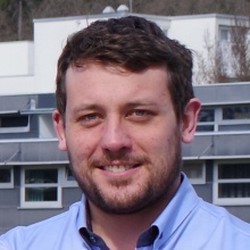}}]{Gabriele Santin} is an Assistant Professor at the Ca' Foscari University of Venice. 
He received his B.Sc., M.Sc., and Ph.D. in Applied Mathematics from the University of Padova, Italy. He was a postdoctoral researched at the 
University of 
Stuttgart, Germany, and a researcher in the Mobile and Social Computing Lab
(MobS) at the Digital Society Center of the Bruno Kessler Foundation in Trento, Italy, and an associated member of the Cluster of Excellence 
Data-Integrated Simulation Science in Stuttgart.
His research interests are at the intersection of machine learning an computational sciences.
\end{IEEEbiography}
\vskip -2\baselineskip plus -1fil
\begin{IEEEbiography}[{\includegraphics[width=1in,height=1.25in,clip,keepaspectratio]{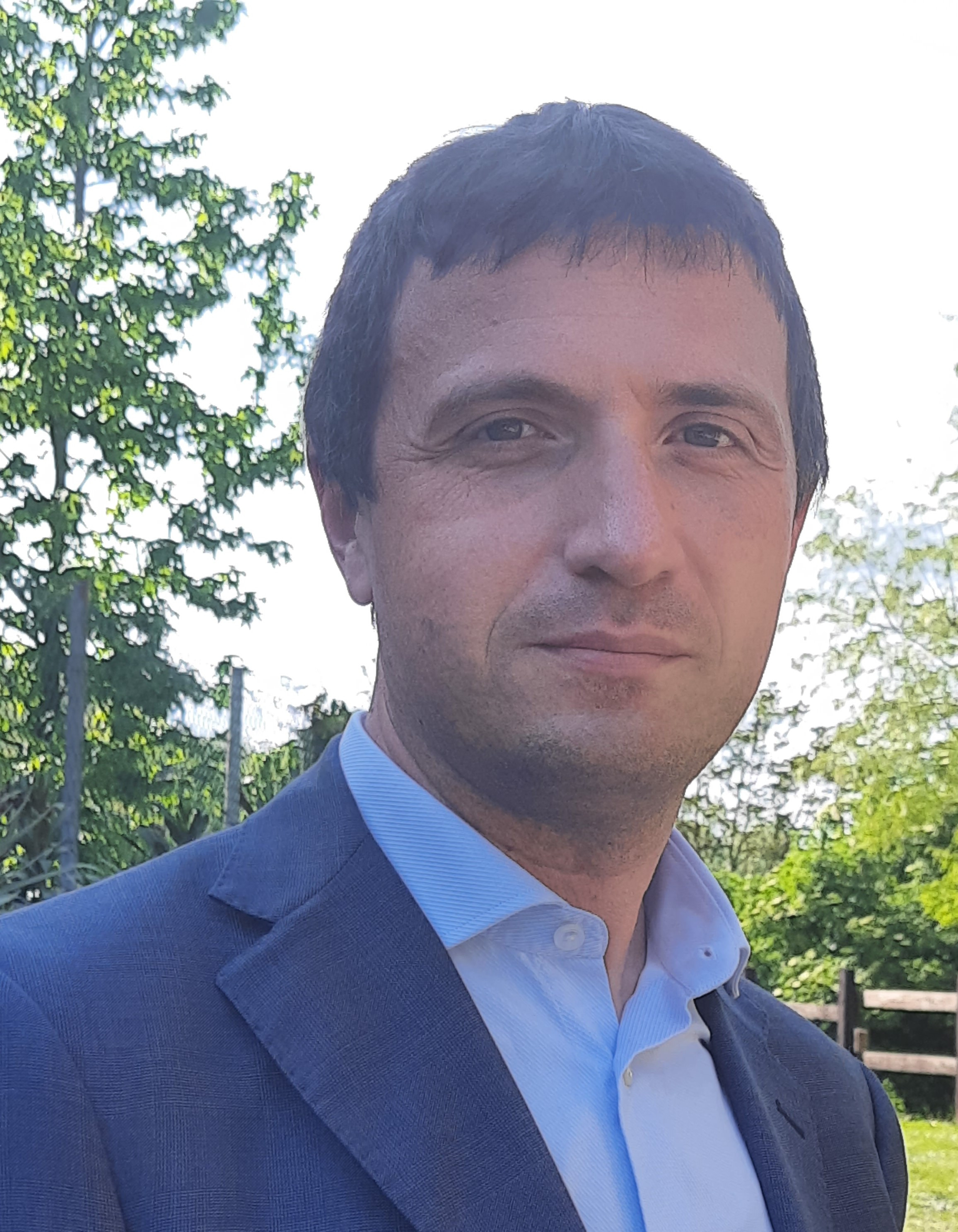}}]{Giorgio Previati}
	is associate professor at Politecnico di Milano, where he got his Ph.D in 2006. His research activity deals in general with machine and vehicle design with particular reference to vehicle and vehicle subsystems testing, structural and multidisciplinary optimization and material modeling. He contributed to the development of a series of test rigs for the measurement of inertia properties and sensors for force and moment measurement. He also worked in the R\&D department of a leading farm tractor industry and in 2010 was visiting researcher at TU Dresden. Prof. Previati is author of over 100 peer reviewed papers, 2 books and 1 patent.
 \end{IEEEbiography}
\vskip -2\baselineskip plus -1fil
 \begin{IEEEbiography}[{\includegraphics[width=1in,height=1.25in,clip,keepaspectratio]{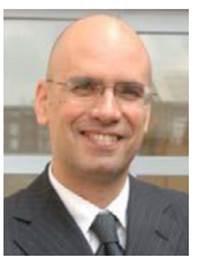}}]{Gianpiero Mastinu}
	is full professor of Road and Off-road Vehicles at Politecnico di Milano. Born in Como on August 4th, 1960. Graduated with honours at the Politecnico di Milano (1984). Doctoral degree in Applied Mechanics (1989). Appointed as Full Professor at TU Delft (2000). Full professor at Politecnico di Milano (2004). Co-founder of the Laboratory for Safety of Transport systems (LaST) of the Politecnico di Milano and coordinator of the Active Safety Division. Founder of the Driving Simulator Laboratory of the Politecnico di Milano. Secretary General of the Cluster Mobility of Lombardy (Government of Lombardy Region). Coordinator of the Scientific Board of the National Mobility Cluster of Italy. Member of the EAG of the FIA. Formerly director of the Master “Vehicle Engineering” organized by University of Modena and Politecnico di Milano, (sponsored by primary automotive companies from Emilia and Lombardy). Author of five books and of about 200 papers on International Journals. His scientific activity has focused on the design, construction, and test of machines, structures, and mechanical systems, with particular reference to ground vehicles. Topics: conceptual, functional and structural design (optimization methods); experimental mechanics; modelling and simulation of machines; road and rail vehicle system engineering. Prof. Mastinu was President of the XIX IAVSD Symposium (Milan, Aug 2005).
 \end{IEEEbiography}
\vskip -2\baselineskip plus -1fil
  \begin{IEEEbiography}[{\includegraphics[width=1in,height=1.25in,clip,keepaspectratio]{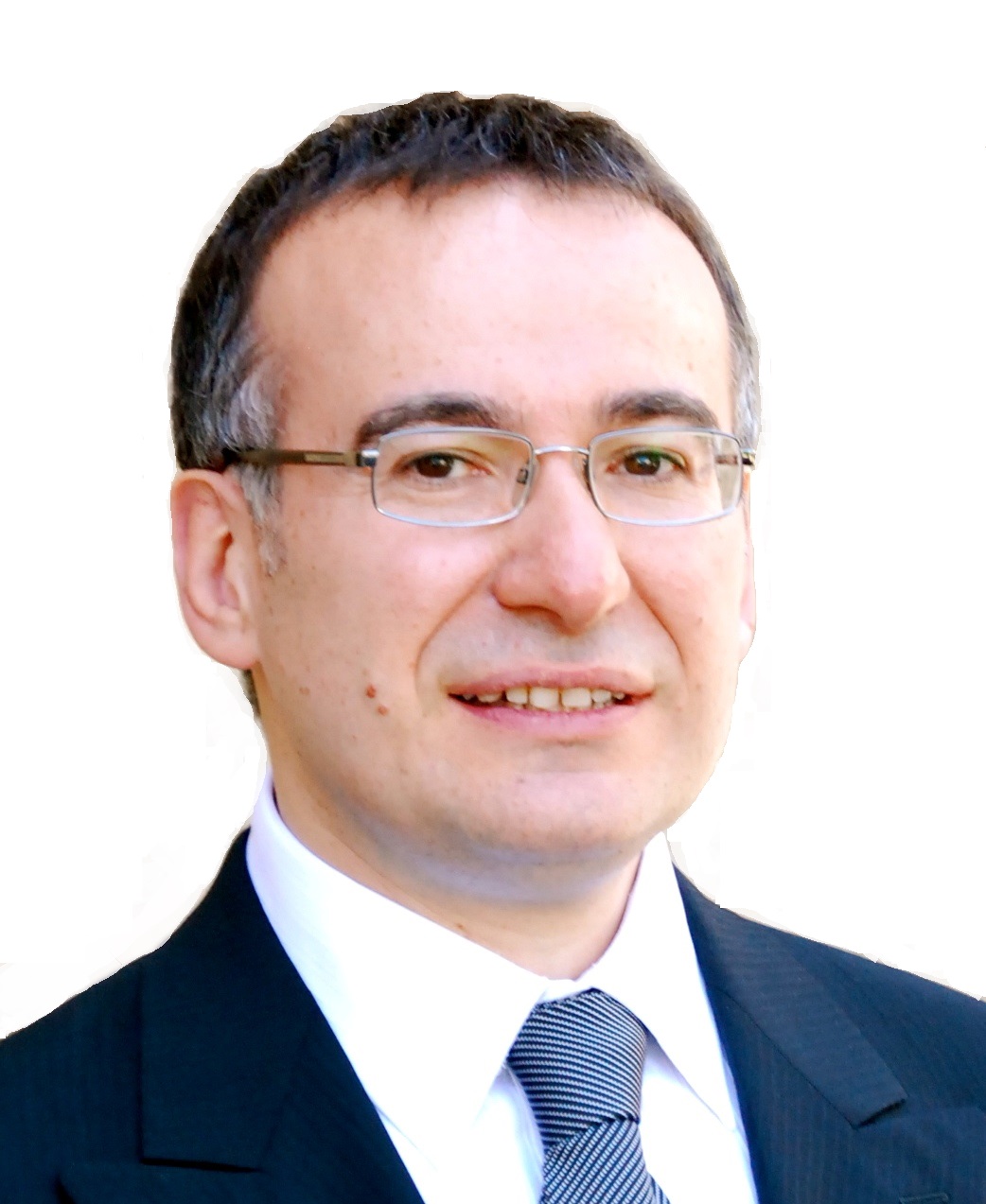}}]{Massimiliano Gobbi}
	received the master’s degree (Hons.) in mechanical engineering in 1994 and the Ph.D. degree in applied mechanics in 1997. In 1998, he was a Visiting Scholar with the Department of Mechanical Engineering, University of California at Berkeley, Berkeley, CA, USA. He is currently a Full Professor with the Politecnico di Milano, Milan, Italy. He has authored more than 200 articles. He is the author of seven international patents and three books.
 \end{IEEEbiography}
 \vskip -2\baselineskip plus -1fil
 \begin{IEEEbiography}[{\includegraphics[width=1in,height=1.25in,clip,keepaspectratio]{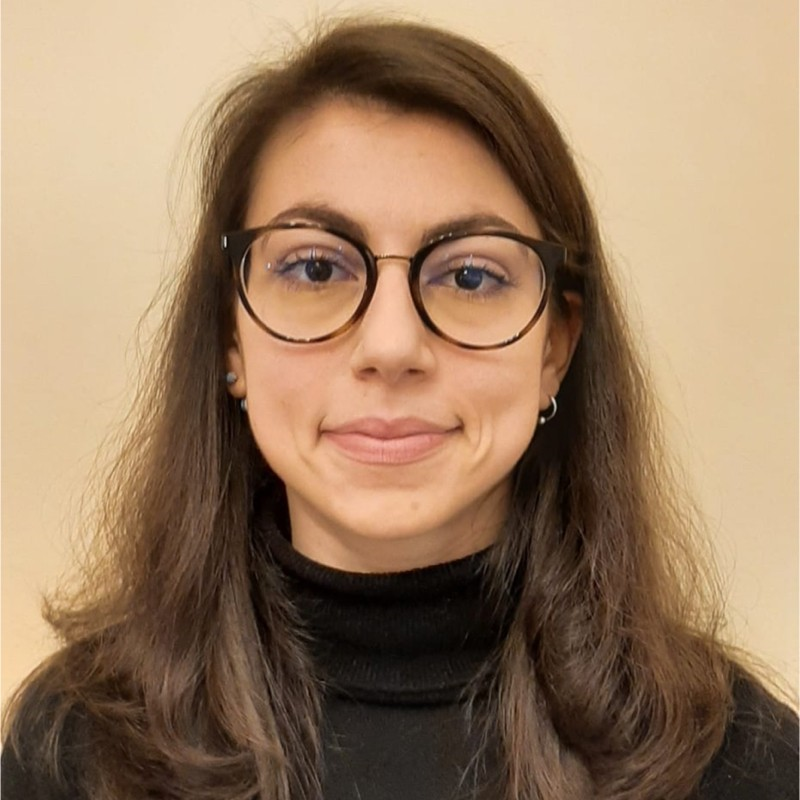}}]{Elena Campi}
	was born in Genoa in 1996. She received her B.Sc. degree in Mechanical Engineering and her M.Sc. in Ground Vehicles at Politecnico di Milano, in 2018 and in 2021 respectively. At the moment, she is a fellowship researcher working at DriSMi, the driving simulator laboratory of Politecnico di Milano. Her current research is focused on traffic simulations with human-in-the-loop and on control algorithms of autonomous and connected vehicles.
 \end{IEEEbiography}
\vskip -2\baselineskip plus -1fil
\begin{IEEEbiography}[{\includegraphics[width=1in,height=1.25in,clip,keepaspectratio]{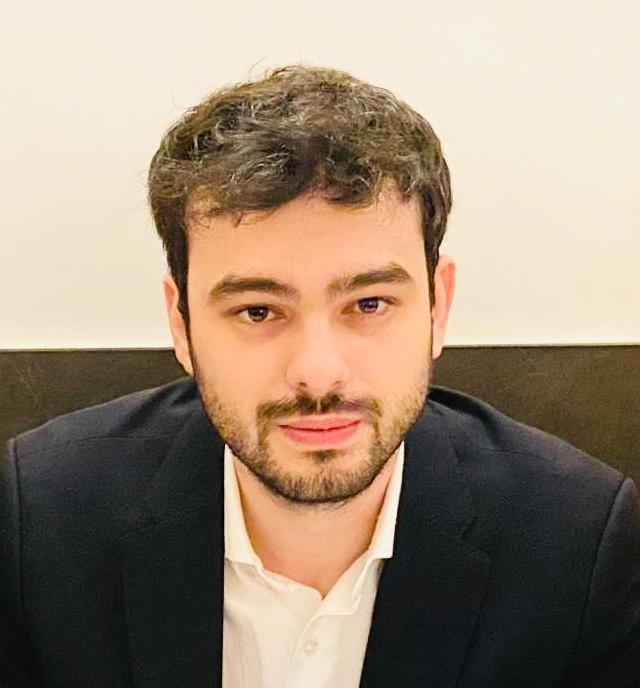}}]{Lorenzo Uccello}
	is researcher at Politecnico di Milano. He received his M.Sc. in Mechanical Engineering, Ground Vehicles, in 2023. He is working at DriSMi, the dynamic driving simulator of Politecnico di Milano. His research focuses on cooperative and automated driving, spanning diverse areas, ranging from communication protocols and simulation to the implementation and validation of digital twins. He is exploring the effect due to the human factor, proactively modifying the simulation environment's performance and leveraging state-of-the-art sensors like EEG, ECG, and SPR with the purpose of analyzing human-machine interaction.	 
    \end{IEEEbiography}
\vskip -2\baselineskip plus -1fil   
\begin{IEEEbiography}[{\includegraphics[width=1in,height=1.25in,clip,keepaspectratio]{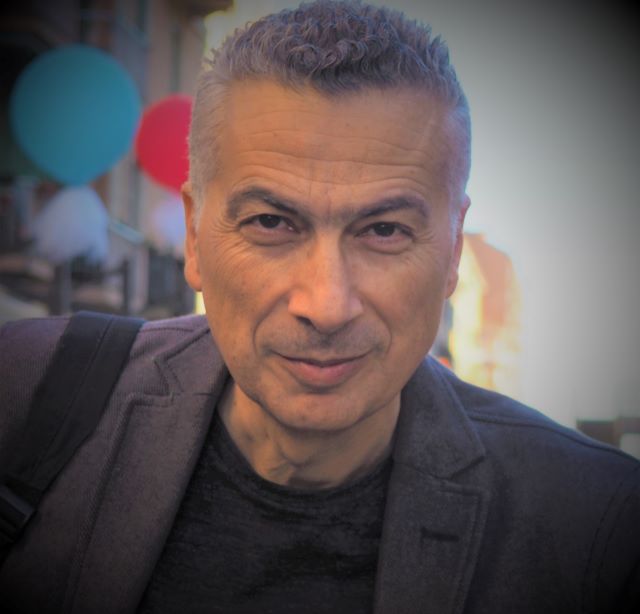}}]{Antonino Albanese} is a Research and Innovation Manager at Italtel S.p.A (Milan, Italy). He has a long experience in the design, development and testing of telecommunication products related to fixed, mobile, and multiservice networks (voice/data/video), distributed on cloud and virtualized infrastructures.
Antonino has management expertise in European collaborative research and innovation projects, being actively involved in projects related to several ICT areas, including telecommunications, cloud, network function virtualization, service defined networking, IoT, Edge computing, AI, and cybersecurity. 
His main research interests include High Performance Computing and HW acceleration technologies that can provide a performance boost in data processing at the Edge Cloud, as well as improve energy and cost efficiency.
\end{IEEEbiography}
\vskip -2\baselineskip plus -1fil
\begin{IEEEbiography}[{\includegraphics[width=1in,height=1.25in,clip,keepaspectratio]{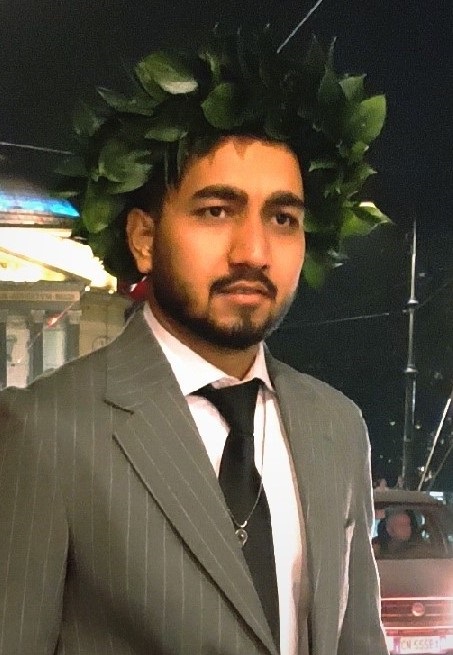}}]{Praveen Zalaya}
	works at Stellantis Turin, Italy, as a Pre-Development Engineer. In 2022, he graduated with an M.Sc. in Autonomous and Connected Vehicle Engineering from Politecnico di Torino. In 2019, he earned a Bachelor of Technology in Mechanical Engineering from Indian Institute of Technology (IIT). He works in the Connected Vehicle area, driving innovation for various cutting-edge connectivity technologies such as 5G, C-V2X, and Satellite communication. He is concentrating his efforts on developing multiple UC's for V2X applications in a simulated environment, merging the ADAS system with enhanced perception capabilities from the V2X system. He also works on application SW development for complete E2E prototypes of various connected car services and safety-related tasks.
    \end{IEEEbiography}
\vskip -2\baselineskip plus -1fil
\begin{IEEEbiography}[{\includegraphics[width=1in,height=1.25in,clip,keepaspectratio]{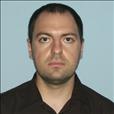}}]{Alessandro Roccasalva}
    works at Stellantis Turin, Italy, as a Testing and validation Engineer. He is graduated in Mechanical Engineering. He has a background in diesel and gasoline testing activities. He performed both development and reliability testing activities (such as engine oil aeration assessment, oil pan level and  blow-by circuit assessment @ tilting bench, max power and piston lubrication assessment). At the moment he works in the Advanced Connectivity team and he is involved in the European project "AI@EDGE" to follow the telematics box module integration with the driving simulator at DriSMi.
    \end{IEEEbiography}
\vskip -2\baselineskip plus -1fil
\begin{IEEEbiography}[{\includegraphics[width=1in,height=1.25in,clip,keepaspectratio]{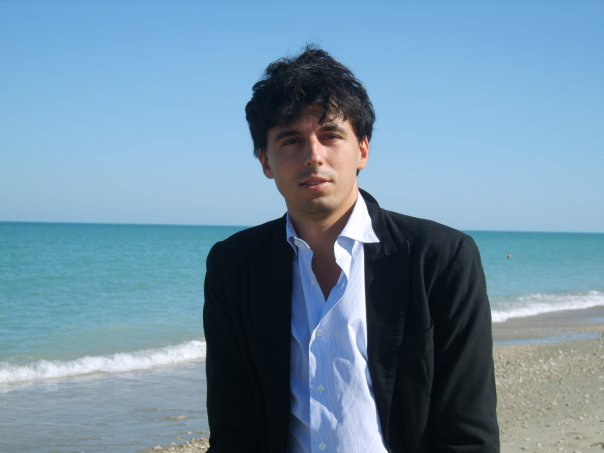}}]{Bruno Lepri} is a senior researcher at Fondazione Bruno Kessler (Trento, Italy) where he leads the Mobile and Social Computing Lab (MobS). He has recently launched the Center for Computational Social Science and Human Dynamics, a joint initiative between Fondazione Bruno Kessler and the University of Trento. Since July 2022, he is the Chief Scientific Officer of Ipazia, a new company active on AI solutions for financial services and energy management. From 2019 to 2022, Bruno was also the Chief AI Scientist of ManpowerGroup where he has collaborated with the global innovation team on AI projects for recruitment and HR management. Bruno is also a senior research affiliate at Data-Pop Alliance, the first think-tank on big data and development co-created by the Harvard Humanitarian Initiative, MIT Media Lab, Overseas Development Institute, and Flowminder. Finally, he has co-founded Profilio, a startup on AI-driven psychometric analysis. In 2010 he won a Marie Curie Cofund postdoc fellowship and he has held a 3-year postdoc position at the MIT Media Lab. He holds a Ph.D. in Computer Science from the University of Trento. His research interests include computational social science, personality computing, network science, and machine learning. His research has received attention from several international press outlets and obtained the 10-year impact award at MUM 2021, the James Chen Annual Award for the best 2016 UMUAI paper, and the best paper award at ACM Ubicomp 2014.
\end{IEEEbiography}
\EOD
\end{document}